\documentclass{IEEEtran4PSCC}

\usepackage[subtle,tracking=normal]{savetrees}

\usepackage{subcaption} 
\usepackage{graphicx} 
\usepackage{multirow}
\usepackage{amsmath}
\usepackage[ruled, lined, linesnumbered, commentsnumbered, longend]{algorithm2e}
\usepackage{xcolor}
\usepackage[utf8]{inputenc}
\usepackage{amsfonts} 
\usepackage{titlesec}
\usepackage{lineno}
\usepackage{color, colortbl}
\usepackage{caption}
\usepackage[hidelinks]{hyperref}
\usepackage{lipsum}

\usepackage{cleveref}[2012/02/15]
\crefformat{footnote}{#2\footnotemark[#1]#3}
\usepackage{colortbl}
\usepackage{cite}

\makeatletter
\let\old@ps@headings\ps@headings
\let\old@ps@IEEEtitlepagestyle\ps@IEEEtitlepagestyle
\def\psccfooter#1{%
    \def\ps@headings{%
        \old@ps@headings%
        \def\@oddfoot{\strut\hfill#1\hfill\strut}%
        \def\@evenfoot{\strut\hfill#1\hfill\strut}%
    }%
    \def\ps@IEEEtitlepagestyle{%
        \old@ps@IEEEtitlepagestyle%
        \def\@oddfoot{\strut\hfill#1\hfill\strut}%
        \def\@evenfoot{\strut\hfill#1\hfill\strut}%
    }%
    \ps@headings%
}
\makeatother

\begin{document}

\title{Graph Isomorphic Networks for Assessing Reliability of the Medium-Voltage Grid}

\author{
    \hspace*{-12pt} \IEEEauthorblockN{Charlotte Cambier van Nooten\IEEEauthorrefmark{1}, Tom van de Poll\IEEEauthorrefmark{2}, Sonja Füllhase\IEEEauthorrefmark{1}, Jacco Heres\IEEEauthorrefmark{2}, Tom Heskes\IEEEauthorrefmark{1}, Yuliya Shapovalova\IEEEauthorrefmark{1}}
    \IEEEauthorblockA{\IEEEauthorrefmark{1} Institute for Computing and Information Sciences, Radboud University Nijmegen, Netherlands \\\{charlotte.cambiervannooten, tom.heskes, yuliya.shapovalova\}@ru.nl, \{sonja\}@fuellhase.de}
    \IEEEauthorblockA{\IEEEauthorrefmark{2} Alliander, Netherlands \\\{tom.van.de.poll, jacco.heres\}@alliander.com}}

\maketitle

\thanksto{\noindent This work is funded by the Dutch Research Council (NWO) ROBUST project (with project number - of the research programme ICAI AI for Energy Grids Lab). The funders played no role in the collection, analysis, interpretation of data or in the writing of the manuscript.}

\begin{abstract}
Ensuring electricity grid reliability becomes increasingly challenging with the shift towards renewable energy and declining conventional capacities. Distribution System Operators (DSOs) aim to achieve grid reliability by verifying the n-1 principle, ensuring continuous operation in case of component failure. Electricity networks’ complex graph-based data holds crucial information for n-1 assessment: graph structure and data about stations/cables. Unlike traditional machine learning methods, Graph Neural Networks (GNNs) directly handle graph-structured data. This paper proposes using Graph Isomorphic Networks (GINs) for n-1 assessments in medium voltage grids. The GIN framework is designed to generalise to unseen grids and utilise graph structure and data about stations/cables. The proposed GIN approach demonstrates faster and more reliable grid assessments than a traditional mathematical optimisation approach, reducing prediction times by approximately a factor of 1000. The findings offer a promising approach to address computational challenges and enhance the reliability and efficiency of energy grid assessments.
\end{abstract}


\section{Introduction}
\begin{figure}
\captionsetup{font=small}
\small
    \centering
    \includegraphics[width=3.5in]{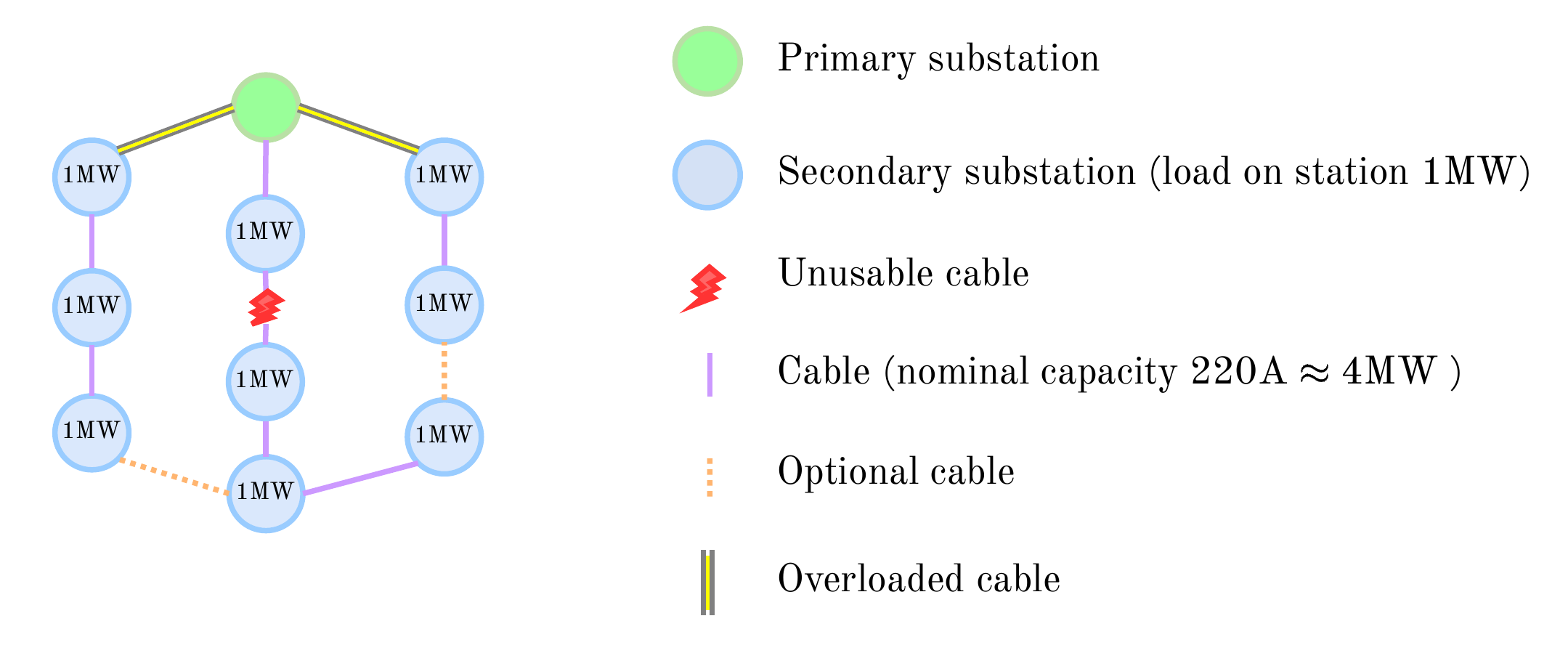}
    \caption{Example medium voltage grid (10.5 kV) distributing power from one primary substation to several secondary substations. It is not possible to reroute the power after an outage in the cable indicated with the red lightning bolt due to capacity limitations in the cables indicated in bold yellow (for simplicity, the reactive power and losses are ignored). In this example, the grid does not satisfy the n-1 property.
    }
    \label{fig:n-1}
\end{figure}

In the modern world, with increased energy usage, diversity of energy sources and sustainability goals, power grid operators face transport capacity limitations and other challenges \cite{fusco_knowledge-_2021}. The energy grid is a complex system. In recent years, there has been increased interest in developing more efficient and reliable methods for managing this critical infrastructure \cite{uddin_next-generation_2023}. An electricity grid operator or Distribution System Operator (DSO) typically desires to ensure that when a line/cable fails in the medium voltage grids, the electricity can be rerouted through other connections such that all consumers still have electricity while at the same time, the currents and voltages are still within the constraints of nominal currents and the maximum allowed voltage deviations. The power system should be designed and operated to remain operational even if any single component fails. For this, the so-called n-1 property is checked, which is crucial for the concept of grid reliability and efficient assessment methods. In practical terms, the n-1 principle means that power system planners and operators must ensure that the grid remains stable and operational despite unexpected events such as equipment failures, overloaded cables, or natural disasters \cite{zhang_mixed-integer_2018}. The n refers to the total number of components in the system, and when n-1 property is satisfied, a failure of 1 component in the grid does not cause a system-wide outage. Figure \ref{fig:n-1} illustrates an example of a grid that does not satisfy the n-1 property, due to overloaded cables when rerouting energy. 

Existing algorithms for checking n-1 property \cite{zhang_mixed-integer_2018, vazquez_reliable_2020, schwerdfeger_approach_2016, fritschy_h_checking_2018} rely on physical constraints and are inefficient in terms of computation time. Frequently used methods belong to the class of mathematical optimisation approaches \cite{fritschy_h_checking_2018, wang2022robust, xue2021dynamic, stock2020operational}. For example, a Dutch DSO deploys a mathematical optimisation approach that iterates through various switching options \cite{fritschy_h_checking_2018, zhang_mixed-integer_2018}. However, this method is often too slow to compute many load scenarios for operating energy grids. Even with only two load scenarios, a single check, e.g. a customer request, can take up to a few hours since, for assessing the n-1 property, calculations should be made for all switching options in the entire graph. Given the goals in energy transition and smart grid solutions, checking the n-1 property reliably and quickly is desirable.

We propose to employ Graph Neural Networks (GNN) \cite{wu_graph_2022, zhang_end--end_2018} to assess the n-1 principle. Graph Neural Networks (GNNs) are a class of deep learning methods designed to perform inference on data naturally represented as graphs. GNNs can be directly applied to graphs and perform node-level \cite{liu_calibration_2022}, edge-level \cite{zhuo_graph_2021} and graph-level \cite{zhang_end--end_2018} prediction tasks. The key design element of GNNs is pairwise message passing, such that graph nodes iteratively update their representations by exchanging information with their neighbours. 
The GNN has several advantages compared to standard algorithms: it helps to tackle the computational burden, directly deals with graph data, and includes topological information from analysed graphs \cite{zhang_end--end_2018}. The currents and voltages highly depend on connected loads and the physical properties of the neighbouring elements of the grid. Using neighbouring elements enables us to make a reliability assessment of whether a grid satisfies the n-1 principle. 

Several GNN architectures have been exploited for graph classification tasks \cite{zhang_end--end_2018, shlomi_graph_2021}. In this paper, we deploy a variation of GNNs, called Graph Isomorphic Networks (GIN), and extend it by including edge features \cite{xu_how_2019, kim_understanding_2020}. GINs modify the message-passing layers to distinguish different graph structures. The GIN uses node features and the adjacency matrix to describe load and topology information. 
By integrating node features and edge features in the process of message passing (updating and aggregation), the impact of topology changes on the nodes is taken into account. Additionally, the proposed approach has the desirable property of generalising to unseen grids. Our approach can significantly reduce the computational burden and help create a more reliable and secure energy grid. 

Related work includes previous studies that have explored various approaches to modelling and analysing the energy grid, including machine learning and optimisation methods \cite{sonja_fullhase_testing_2020, schwerdfeger_approach_2016, qian_n-1_2022, vazquez_reliable_2020}. However, to the best of our knowledge, this is the first study to apply GNNs to the n-1 principle of the energy grid in order to maximise the grid's reliability of unseen grids. 

\section{The proposed methodology} 
We describe the proposed methodology in the following order. First, the necessary notation is introduced for describing the energy grid as a graph. Second, we discuss data augmentation using existing real-world topologies. Finally, we introduce Graph Isomorphic Networks (GINs) for the problem of n-1 graph classification.

\subsection{Energy grid as a graph}
We define the energy grid as an undirected graph $G=(V,E)$, where $V$ the set of nodes/vertices represents the substations, and $E$ the set of edges represents the cables of the energy grid. Here, $V$ and $E$ are finite. An edge $e$ connects nodes $u$ and $v$ if they are adjacent (neighbours). The following notation is used throughout the paper:

\begin{itemize}
    \item $|V|$, number of nodes.
    \item $|E|$, number of edges.
    \item $\mathbb{G}$, input set of graphs, where $\mathbb{G} = \{ G_1, G_2, \dots, G_n\}$.
    \item $Y$, input set of graph labels.
    \item ${F}^V$, input set of all node features for a specific graph, ${F}^V = \{F^V_1, . . . , F^V_{|V|}\}$.
    \item ${F}^E$, input set of all edge features for a specific graph, ${F}^E = \{F^E_1, . . . , F^E_{|E|}\}$.
    \item $F^V_v \in \mathbb{R}^B$, node feature matrix of $v$-th node ($v=1,\dots,|V|$), where node features are in $B$-dimensions. 
    \item $F^E_e \in \mathbb{R}^P$, edge feature matrix of $e$-th edge ($e=1,\dots,|E|$), where edge features are in $P$-dimensions (when there is no edge, $F^E_e = F^E_{(u,v)}=0$).
    \item $\mathcal{N}(v)$, neighbouring nodes of node $v$, $v=1,\dots,|V|$.
    \item ${X}$, all input data, ${X}_i = \{G_i, {F}^V, {F}^E\}$, containing the graph structure and corresponding node and edge features. The features are different between graphs, we omit this for simplicity of notation.
    \item $D = \{(X_1, y_1), (X_2, y_2), \dots, (X_n, y_n) \}$, set containing input data and labelled graphs.
    \item $h_v^{(k)}$, hidden state of node $v$ in the $k$-th layer.
    \item $g_e^{(k)}$, hidden state of edge $e$ in the $k$-th layer.
\end{itemize}

\subsection{Data augmentation: graph structure, node and edge features} \label{subsec:topo}
In this paper, we use real grid data (obtained from Alliander, DSO in the Netherlands) together with augmented data to assess whether a grid satisfies n-1 property \cite{fritschy_h_checking_2018}. The dataset without augmentation contains 32000 samples; each sample includes a graph structure and corresponding features. The data were generated using four fixed real grid structures in different locations in the Netherlands, while the node features ($F^{V}$) and edge features ($F^{E}$) were generated using the grid feature generation method based on load forecasting and load flow\cite{van2017andes, fritschy_h_checking_2018}. Each sample, containing a graph structure and features, is associated with a binary input label (n-1 or not n-1). Before augmentation, for each location, half of the samples were labelled as n-1 and half as not n-1. These data are summarised in Table \ref{tab:stat_data}. 

We employ data augmentation through topology adaptations to enhance dataset diversity and promote the model's ability to generalise to unseen grids. The four real grid structures are used as a base for the augmentations. For all of the original 32000 samples each sample is augmented once by changing the grid structure and corresponding features, resulting in a full dataset of 64000 samples. The complete dataset is summarised in Table \ref{tab:stat_data_topo} and contains 50\% of graphs labelled as n-1 and 50\% labelled as not n-1. 

The general procedure for simulating new grid structures can be described as follows. We start with graph $G=(V, E)$ and choose candidate nodes for adaptations. If the graph is labelled as n-1, then some candidate nodes are added (as this does not change the label), and if the graph is not n-1, then some candidate nodes are removed from the graph structure. The number of nodes to be added/removed from the graph is chosen randomly from the candidate set of nodes. If the nodes have been removed from the graph, the features do not change. However, if the nodes are added to the graph, the features for the added nodes (and corresponding edges) are recomputed. For this, we define a neighbourhood around the newly added nodes reachable within two connections (2-hops away) and compute new features by averaging all the feature values of neighbouring elements. Features that cannot be averaged over (e.g., node degree), are recomputed for all nodes/edges in the graph.

\begin{figure*}[]
\captionsetup{font=small}
\small
    \centering
    \includegraphics[width=4.5in]{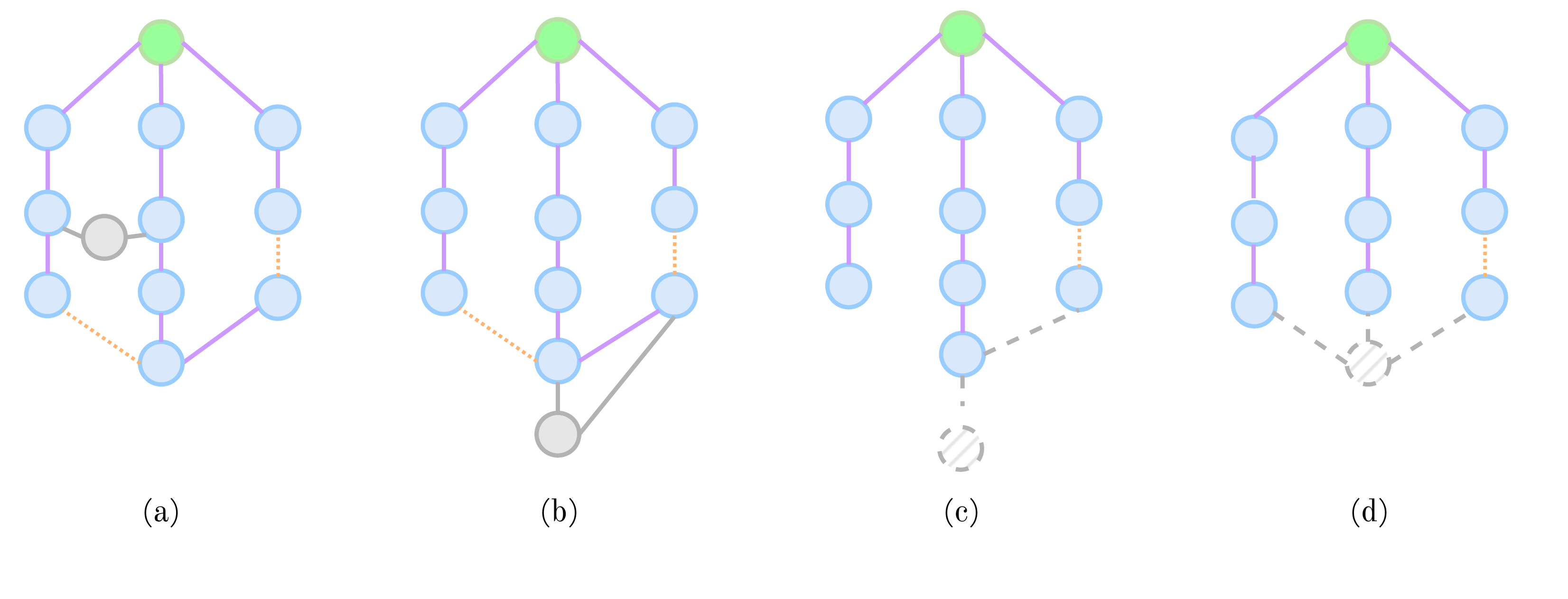}
    \caption{Examples of data augmentation.
    The purple and orange lines illustrate the starting topology. a) topology with the n-1 label, adding nodes (grey) while keeping the total load constant, will keep the label. b) topology with the n-1 label, adding nodes (grey) will not change the label. c) topology with the not n-1 label, removing nodes (grey) will keep the label (not n-1). d) topology with the n-1 label, removing nodes (grey) will break the n-1 property.}
    \label{fig:topochange}
\end{figure*}

Figure \ref{fig:topochange} illustrates the intuition behind the augmentation of the grid structures. Figures \ref{fig:topochange}a, \ref{fig:topochange}b, and \ref{fig:topochange}c illustrate the type of augmentation that preserves the grid label. Specifically, Figures \ref{fig:topochange}a and \ref{fig:topochange}b are labelled as n-1 and adding nodes preserves the label. Similarly, Figure \ref{fig:topochange}c is not n-1 and removing nodes preserves the not n-1 label. Finally, we avoid situations where the augmentation may lead to the change of the label. Figure \ref{fig:topochange}d  illustrates an example with the graph labelled as n-1, and removing nodes leads to a different label, i.e. not n-1. 

\begin{table}[]
\captionsetup{font=small}
\small
\centering
\caption{Node and edge features used for n-1 graph classification with Graph Neural Networks.}
\label{tab:feat_table}
\begin{tabular}{ll}
\hline
Node features (stations)            & Edge features (cables)                                                                   \\ \hline
Power consumption & Impedance \\ (or generation on load of node)                 & Nominal current      \\
Voltage of (radial) initial state & Nominal current                                                                                \\
Voltage of closed state  & Current of initial (radial) state                                                                           \\
Node degree                         & Current of closed state                                                                      \\ \hline 
\end{tabular}
\end{table}

\begin{table}[]
    \captionsetup{font=small}
\small
    \caption{The summary of graph datasets. $\#$Samples denotes the number of samples (graphs), $\#$Nodes denotes the number of nodes among the graphs, and $\#$Edges denotes the number of edges. a) Original dataset; b) Augmented dataset, nodes and edges are given in average and standard deviation, in terms of number removed/added.}
    \label{tab:stat_data}
    \begin{subtable}{0.5\textwidth}
      \centering
        \caption{}
        \begin{tabular}{rrrr}
        \hline
        Location & $\#$Nodes & $\#$Edges & $\#$Samples \\ \hline
        1        & 394   & 431   & 10000   \\
        2        & 57    & 64    & 10000   \\
        3        & 73    & 77    & 10000   \\
        4        & 878   & 1001  & 2000    \\ \hline
        \end{tabular}
        \label{tab:stat_data}
    \end{subtable}%
     \hfill
    \begin{subtable}{.5\textwidth}
      \centering
        \caption{}
        \begin{tabular}{rrrr}
        \hline
        Location & $\#$Nodes & $\#$Edges & $\#$Samples \\ \hline
        1        & 347 $\pm$ 101 & 388 $\pm$ 120  & 10000   \\
        2        & 43 $\pm$ 21  & 49 $\pm$ 23  & 10000   \\
        3        & 51 $\pm$ 39  & 59  $\pm$ 41 & 10000   \\
        4        & 790 $\pm$ 211 & 823 $\pm$ 334 & 2000    \\ \hline
        \end{tabular}
        \label{tab:stat_data_topo}
    \end{subtable} 
\end{table}

\subsection{Graph reliability classification}
The goal of GNNs for the graph-supervised classification task is to learn a non-linear mapping $f$ from the input data to the output space
\begin{equation}\label{eq:g}
    f: X \mapsto Y,
\end{equation}
where $X$ is the input data, and $Y$ is a set of graph labels. The input data ($X$) contains graph structures ($\mathbb{G}$), node features ($F^V$) and edge features ($F^E$). Furthermore, we have input data with labelled graphs $D = \{(X_1, y_1), (X_2, y_2), \dots, (X_n, y_n) \}$. We are interested in assessing whether graph $X_{n+1}$ satisfies the n-1 property.

\subsection{Graph Neural Networks}
Graph Neural Networks (GNNs) \cite{zhang_end--end_2018} are a class of neural networks designed to operate on graph-structured data. A GNN can be defined as a function $f$ that takes a graph $G=(V, E)$ and a set of node features ${F}^V = \{F^V_1, . . . , F^V_{|V|}\}$, edge features ${F} ^E = \{F^E_1, . . . , F^E_{|E|}\}$, and produces a set of output labels $Y = \{y_1, \dots, y_n\}$, where each $y_i$ is a label of the graph structure. The output labels $y_i$ for each graph are computed using a recursive message-passing procedure that operates on the local neighbourhood of nodes and edges. In a GNN, nodes receive messages from their neighbours, which are combined with their own features and neighbouring edge features. This process is repeated for a fixed number of iterations, and the final node features are used to compute the output features on graph-level \cite{zhou_graph_2020}. The message-passing procedure in a GNN is implemented using a neural network module to aggregate the features of neighbouring nodes and combine them with the features of the central node. The network adapts to the input graph through learned layer weights, and spatial relations are based on combining and aggregation.

\subsubsection{Graph Isomorphic Layers}
Typically, in the standard GNN, non-injective aggregation functions (mean, max or sum) are used in the mapping function $f$ \eqref{eq:g} \cite{xu_how_2019}. Due to the use of non-injective aggregation functions, standard GNNs fail to generalise to unseen graph structures, which makes the expressive power of such GNN architectures limited. We overcome this problem by using a Graph Isomorphic Network (GIN) \cite{xu_how_2019}, which can learn the injective aggregation. This makes the model as powerful as the Weisfeiler-Lehman (WL) test \cite{shervashidze_weisfeiler-lehman_2011} for graph classification tasks.

\begin{figure}
\captionsetup{font=small}
\small
    \centering
       \includegraphics[width=3.0in]{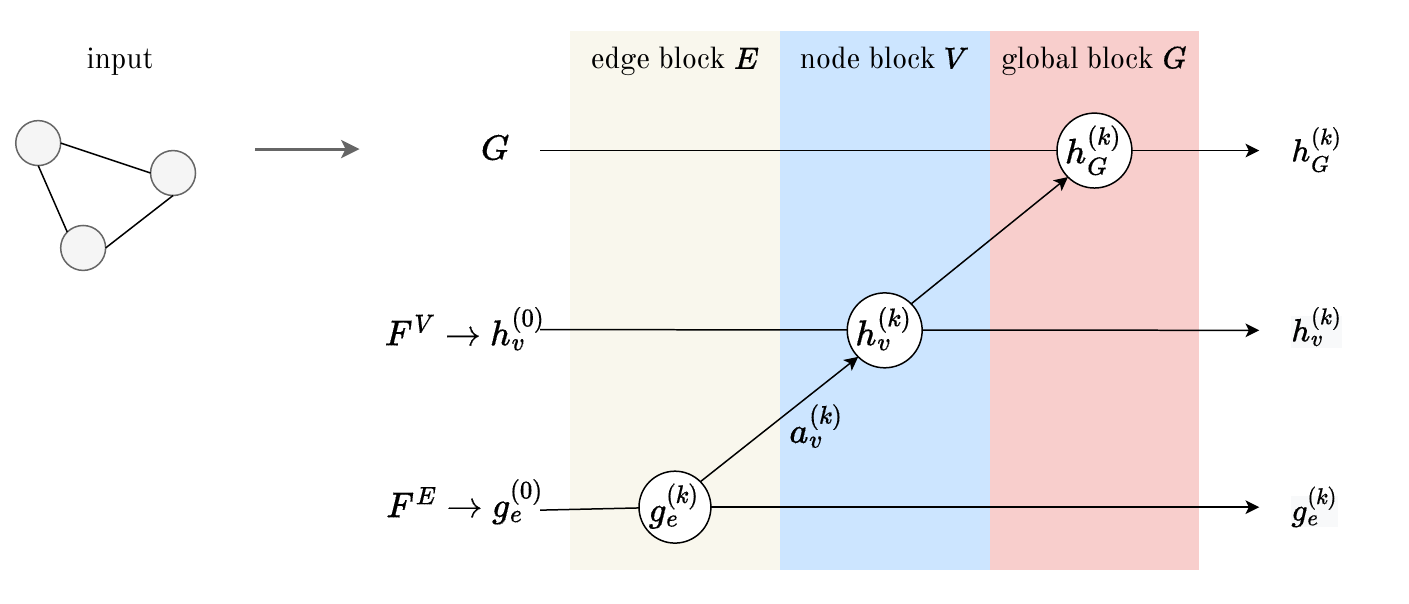}
    \caption{Overview of a single GIN layer, example includes as input the first layer of embeddings ($k=0$), and results in the second layer of embeddings ($k=1$). Updating embeddings involves three blocks: edge-level, node-level, and graph-level.}
    \label{fig:n-1_GIN}
\end{figure}
We proceed by describing the GIN architecture with the overall structure of a GIN layer illustrated in Figure \ref{fig:n-1_GIN}\footnote{\label{note1} More detailed illustrations of each block in Figure \ref{fig:n-1_GIN} and code are available at \url{https://github.com/charlottecvn/GINenergygrids}.}. In energy grid data, raw node features $F^{V}_{u}$, $u=1,\dots,|V|$, and edge features $F_{e}^{E}$, $e=1,\dots,|E|$, are vectors. Embeddings are computed for all nodes and edges, resulting in feature matrices of $|V| \times 16$ and $|E| \times 16$, respectively. To provide input features to the GIN, we first learn embeddings from the raw features as
\begin{equation} \label{eq:embd_node}
    h_{v}^{(0)} = \sum_{u \in \mathcal{N}(v)} MLP_1(F^V_u),
\end{equation}

\begin{equation} \label{eq:embd_edge}
    g_{e}^{(0)} = MLP_2(F^E_e),
\end{equation}
where $MLP_1$ and $MLP_2$ are multi-layer perceptrons with non-linearity used for the embedding of node and edge features correspondingly, $u \in \mathcal{N}(v)$ represents the neighbouring elements of target node $v$ and $F^E_{e}$ the edge features associated with edge $e$.  Each $MLP$  has the same amount of layers, parameters and learnable parameters. The weights of the different $MLP$s are optimised separately. The embeddings $h_{v}^{(0)}$ are computed for each node, $v=1,\dots, |V|$ and the embeddings $g_{e}^{(0)}$ are computed for each edge, $e=1,\dots,|E|$.

The update function for the edge features collects the edge embeddings for layer $k$
\begin{equation}\label{eq:update_edge}
    g_e^{(k)} = MLP_3^{(k)} ((1+\epsilon_1^{(k)})\cdot g_e^{(k-1)}),
\end{equation}
where $\epsilon_1^{(k)}$ is a learnable parameter, $MLP_3^{(k)}$ the multi-layer perceptron for the learnable edge features, and $g_{e}^{(k-1)}$ the edge embeddings surrounding edge $e$. $g_{e}^{(k)}$ is computed for each edge, $e=1,\dots, |E|$, resulting in a feature matrix of size $|E| \times 16$. 

Next, the node embeddings $h_v^{(k)}$ of the current $k$-th layer are computed by using aggregation and combined functions. The aggregation function collects node and edge embeddings of the neighbourhood nodes to extract aggregated feature vectors $a_v^{(k)}$ for the layer $k$ 
\begin{equation}
    a_v^{(k)} = \sum_{u \in \mathcal{N}(v)} ReLU (h_u^{(k-1)}+g_{(v,u)}^{(k-1)})),
\end{equation}
where the aggregated embedding $a_v^{(k)}$ is computed for each node in the graph ($|V|$ times), resulting in a feature matrix of size $|V| \times 16$. To compute $a_v^{(k)}$ we take the target node $v$ and sum over all neighbouring nodes $u$ of $v$, $u \in \mathcal{N}(v)$. 

The combine function combines the previous node embeddings $h_v^{(k-1)}$ with aggregated node embeddings $a_v^{(k)}$ to output the node embeddings of the current $k$-th layer 

\begin{equation} \label{eq:update_node}
    h_v^{(k)} = MLP_4^{(k)} ((1+\epsilon_2^{(k)})\cdot h_v^{(k-1)} + a_v^{(k)}),
\end{equation}
where $h_v^{(k)}$ is computed for each node in the graph ($|V|$ times), resulting in a feature matrix of size $|V| \times 16$,  $MLP_4^{(k)}$ is the multi-layer perceptron for the learnable node features, and $\epsilon_2^{(k)}$ is a learnable parameter. After $K$ iterations of aggregation, the node representation captures the structural information within its $K$-hop network neighbourhood.  

In the graph-level readout, the node embeddings of every layer are aggregated and then concatenated to obtain the final graph embedding $h_G$ for the layer $k$ 
\begin{equation}
    h_G^{(k)} = READOUT ( \{ h_v^{(k)}|v \in G\}),
\end{equation}
where $v$ denotes the target node. The aggregation of nodes to global values is done by global sum-pooling over vectorial node representations. 
The embedding for graph $G$ at layer $k$, is given by the sum-pooling over the multi-set obtained by $h_v^{(k)}, \forall v \in G$ 
\begin{equation}
    h_G^{(k)} = POOL_{sum} ( \{ h_v^{(k)}|v \in G\}) = \sum_{v\in G} h_v^{(k)}.
\end{equation}
The final representation of graph $G$ concatenates the representations of all $K$ embeddings
\begin{equation}
    h_G = CONCAT (\{h_G^{(k)}\} | k=0,1,...,K-1),
\end{equation}
where $K$ is the number of layers. Final labels are predicted using a global $MLP$ (label prediction) and sigmoid ($\sigma$) function.

\subsubsection{$K$-hop neighbourhood}\label{subsec:khop} 
The $K$-hop neighbourhood contains all nodes and edges at most $K$ hops away from node $i$ in the graph. The amount of layers in a GNN is related to the neighbourhood that we want to explore. To integrate information and complex patterns from graphs, the amount of layers in a GNN is set to $K$. In a MV grid, a graph consists of nodes which can be primary or secondary substations (OS/RS/SS) or distribution substations (MSR). The grids are constructed using several OS, each routing to several MSR. The average route length of travelling from an OS to an end MSR is associated with the number of members in the neighbourhood we want to explore. Therefore, we define a $K$-hop neighbourhood, where $K$ is the average route length. Across all samples in the full dataset, the minimum route length is 11, the average is 16, and the maximum route length is 23. Therefore, we expect the optimal $K$ to be around 16, but explore different scenarios between 5 and 20. 

\section{Experiments}

\subsection{Training Details}
All experiments are conducted using PyTorch 1.13.1 and PyTorch Geometric with the GIN framework\cref{note1}. The test set is sampled exclusively from the real grid data and contains 20\% of the real data. The training and validation data contain the real grid data and augmented data, as discussed in Section \ref{subsec:topo}. One-dimensional batch normalisation is applied after each network layer. The parameters $\epsilon_1^{(k)}$  and  $\epsilon_2^{(k)}$ (see \eqref{eq:update_edge} and \eqref{eq:update_node}) are learned by gradient descent. 
The embeddings are created using two-layer $MLP$s. For the individual network configurations, 15 GIN layers (excluding the input layer) are applied. We use the Adam optimiser with a $1 \times 10^{-4}$ learning rate.

\subsection{Mathematical Optimisation Approach}
For a realistic performance evaluation, we compare the proposed GIN framework with a mathematical optimisation approach \cite{ fritschy_h_checking_2018, zhang_mixed-integer_2018}, which is a typical class of methods used by DSOs. The method is based on specifying an objective function subject to many constraints and searching for switching power options for all possible cable failures. If such switching options exist, then the grid is considered to be n-1. However, the mathematical optimisation approach is not an exhausting search for switch options, so it is not guaranteed that a switch option will be found when one exists. The approach is non-trivial, so we refer to \cite{fritschy_h_checking_2018, zhang_mixed-integer_2018, vazquez_reliable_2020, mei_mixed_2023} for more details.

\subsection{Performance Evaluation Metrics} 
We use standard evaluation metrics for the binary classification problem with balanced data: accuracy and the area under the receiver operating characteristic curve (AUC). 

\subsubsection{Accuracy} The accuracy measures the overall performance of the classification model, the ratio of the number of correct classifications to the total number of correct or incorrect classifications
\begin{align}
    \text{accuracy } = \frac{1}{N}\sum_i^N 1(y_i = \hat{y}_i),\label{eq:line}
\end{align}
where $y$ is a vector of target values, and $\hat{y}$ is a vector of predictions.

\subsubsection{The area under the curve (AUC)} The area under the receiver operating characteristic curve (AUC) is often used to evaluate performance in binary classification problems. The AUC score summarises the receiver operating characteristic curve (ROC) into a single number that describes the performance of a model for multiple thresholds simultaneously. Notably, an AUC score of 1 is perfect, and an AUC score of 0.5 corresponds to random guessing. 

\section{Results}
We evaluate and illustrate the proposed framework in several ways. First, we assess the predictive performance of GIN for the n-1 classification problem, including their generalisability to unseen grids. Second, we investigate which graph-related features contribute to the predictive performance. Finally, we compare the proposed approach to a mathematical optimisation approach.

\subsection{Graph Isomorphic Networks (GIN)}
We evaluate the performance of GINs for the n-1 classification problem with several experimental setups. First, we consider different train/test data splits of four different graph structures in reality corresponding to different grid locations. In the first experimental setup (Experiments I) the models are trained and tested on the same single location. In the second set of experiments (Experiments II) all locations are used during training and testing. The final set of experiments (Experiments III) evaluates the performance on unseen grids: the predictions are assessed using a leave-one-out (LOO) approach where three grid structures are used for training and one for testing. Additionally, we evaluate how the performance is affected by data augmentation (with or without), the number of layers ($K$-hop neighbourhood span: 5, 10, 15 and 20) and the choice of global pooling function (sum, mean, max). Table \ref{tab:auc_single} presents the performance results in terms of AUC, and Table \ref{tab:accuracy} (see Appendix A) presents accuracy.

\begin{table*}[]
\captionsetup{font=small}
\small
\centering
\caption{Comparisons of graph classification performance of GIN in terms of AUC of different settings on three experiment sets (single locations, all locations and unseen grids (leave-one-out (LOO))). Here, we implement four versions of GIN with different amounts of layers, three with different global pooling functions, and a version without data augmentation. Best AUC is highlighted in bold.}
\label{tab:auc_single}
\begin{tabular}{l|llll!{\color{gray}\vrule}l!{\color{gray}\vrule}llll|ll}
\hline
 & \multicolumn{4}{c!{\color{gray}\vrule}}{Experiments I} & \multicolumn{1}{c!{\color{gray}\vrule}}{Experiments II} & \multicolumn{4}{c|}{Experiments III} &  &  \\ 
\multirow{-2}{*}{Model} & \multicolumn{1}{c}{1} & \multicolumn{1}{c}{2}  & \multicolumn{1}{c}{3}  & \multicolumn{1}{c !{\color{gray}\vrule}}{4} & \multicolumn{1}{c!{\color{gray}\vrule}}{1-4} & \multicolumn{1}{c}{LOO 1} & \multicolumn{1}{c}{LOO 2}  & \multicolumn{1}{c}{LOO 3}  & \multicolumn{1}{c|}{LOO 4} & \multirow{-2}{*}{\begin{tabular}[c]{@{}l@{}}Data \\ Augmentation\end{tabular}} & \multirow{-2}{*}{\begin{tabular}[c]{@{}l@{}}Global \\ Pooling\end{tabular}} \\ \hline
        GIN-5 & 0.68 & 0.71 & 0.72 & 0.67 & 0.78 & 0.53 & 0.57 & 0.58 & 0.51 & True & Sum  \\ 
        \rowcolor[HTML]{EFEFEF} GIN-10 & 0.81 & 0.81 & 0.83 & 0.79 & 0.91 & 0.71 & 0.67 & 0.66 & 0.69 & True & Sum    \\ 
        GIN-15 & \textbf{0.94} & \textbf{0.95}&\textbf{0.95} &\textbf{0.94} & \textbf{0.96}& \textbf{0.92} & \textbf{0.84}& \textbf{0.87} &\textbf{0.90}& True & Sum    \\ 
        \rowcolor[HTML]{EFEFEF} GIN-20 & 0.89 & 0.90 & 0.90 & 0.89 & 0.96 & 0.88 & 0.80 & 0.83 & 0.87 & True & Sum    \\ 
        GIN-15 & 0.84 & 0.82 & 0.80 & 0.82 & 0.87 & 0.61 & 0.54 & 0.55 & 0.62 & False & Sum    \\ 
        \rowcolor[HTML]{EFEFEF} GIN-15 & 0.93 & 0.91 & 0.90 & 0.92 & 0.95 & 0.88 & 0.86 & 0.84 & 0.89 & True & Max    \\ 
        GIN-15 & 0.85 & 0.83 & 0.86 & 0.86 & 0.94 & 0.82 & 0.79 & 0.77 & 0.80 & True & Mean    \\  \hline

\end{tabular}
\end{table*}

\begin{figure}
    \captionsetup{font=small}
    \small
     \centering
     \begin{subfigure}[b]{1.7in}
         \centering
         \includegraphics[width=1.6in]{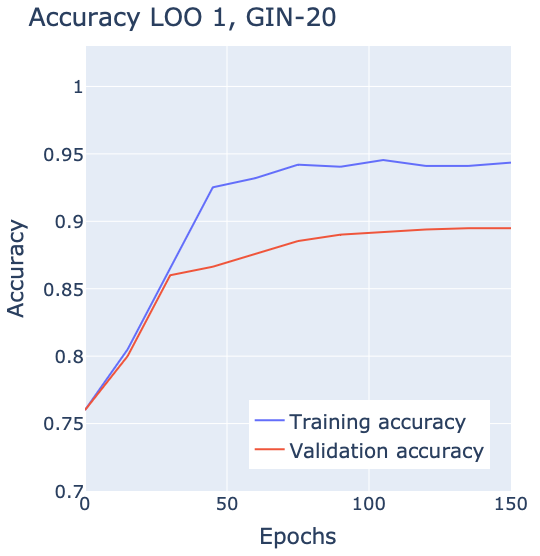}
         \caption{}
         \label{fig:gin20}
     \end{subfigure}
     \hfill
     \begin{subfigure}[b]{1.7in}
         \centering
         \includegraphics[width=1.65in]{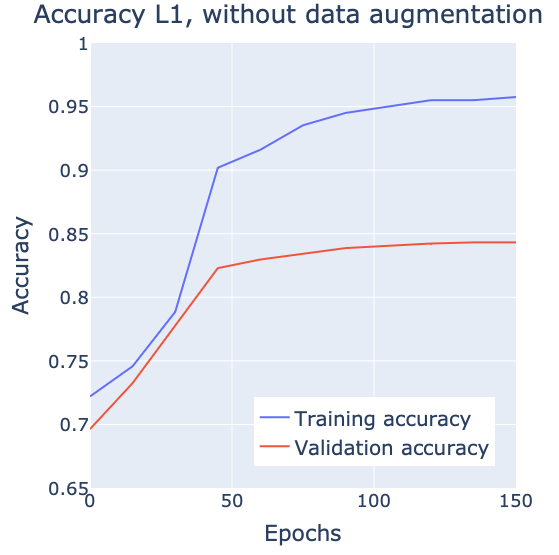}
         \caption{}
         \label{fig:gin15}
     \end{subfigure}
        \caption{a) Comparisons of graph classification performance of GIN-20 in terms of accuracy for LOO-1 (leave-one-out (LOO)). b) Comparisons of graph classification performance of deep GIN-15 in terms of accuracy for location 1 without data augmentation.}
        \label{fig:gin_acc}
\end{figure}

We compare the classification performance given the number of layers used in the GIN framework with 5, 10, 15 and 20 to find the optimal number of layers for graph-level classification. For all different test/training settings (Experiments I, II and III), using a higher number of layers yields higher performance in terms of AUC. This can be explained by the more extensive reach of neighbourhoods that can be explored using more layers. For example, using only five layers (GIN-5) for the leave-one-out experiments (Experiments III) gives scores that are slightly better than random guessing; AUC is 0.53, 0.57, 0.58, and 0.51 for the LOO locations 1-4, correspondingly. But using fifteen layers (GIN-15) leads to corresponding AUCs of 0.92, 0.84, 0.87 and 0.9. However, at some point, using more layers leads to overfitting, as shown in Figure~\ref{fig:gin20}, in which validation accuracy is lower compared to the training accuracy for leave-one-out experiments with GIN-20. 

Furthermore, overfitting is also observed for the single train/test location experiments in the case of no data augmentation, which is illustrated in Figure~\ref{fig:gin15}. Overall, we observe from Tables~\ref{tab:auc_single} and~\ref{tab:accuracy} (see Table~\ref{tab:accuracy} in Appendix A) that experiments without data augmentation (set to False) give lower accuracy and AUC compared to those with data augmentation. For example, the performance for LOO locations 2 and 3 is close to random guessing (GIN-15 without data augmentation for LOO 2 and LOO 3 gives an AUC of 0.54 and 0.55, respectively), while for the experiments with data augmentation, the AUC is high (0.84 and 0.87, respectively). 

Focusing on the Experiments III results in Table \ref{tab:auc_single}, we observe that the performance of GIN-15 to unseen grids gives good discriminatory abilities. The performance in sparse-grid regimes (locations 2 and 3) for GIN-15 is reasonable ($0.87$ AUC). When testing on sparse samples (in terms of the number of nodes and edges), the training set should represent the sparsity of some samples. Testing on larger samples and training on smaller samples results in better performance. As expected, using all four grid structures for test/train sets (Experiments II) results in a higher AUC as the model can better capture the various graph structures.  
Finally, we compare the performance with different global pooling functions. While GIN-15 with sum-pooling and GIN with max-pooling perform similarly on leave-one-out locations, GIN-15 with sum-pooling exhibits higher accuracy (Table~\ref{tab:accuracy} in Appendix A) and AUC (Table~\ref{tab:auc_single}) than max-pooling on augmented data.

\subsubsection{Permutation Feature Importance}
\begin{figure*}[]
     \captionsetup{font=small}
     \small
     \centering
     \begin{subfigure}[b]{3.5in}
         \centering
         \includegraphics[width=2.3in]{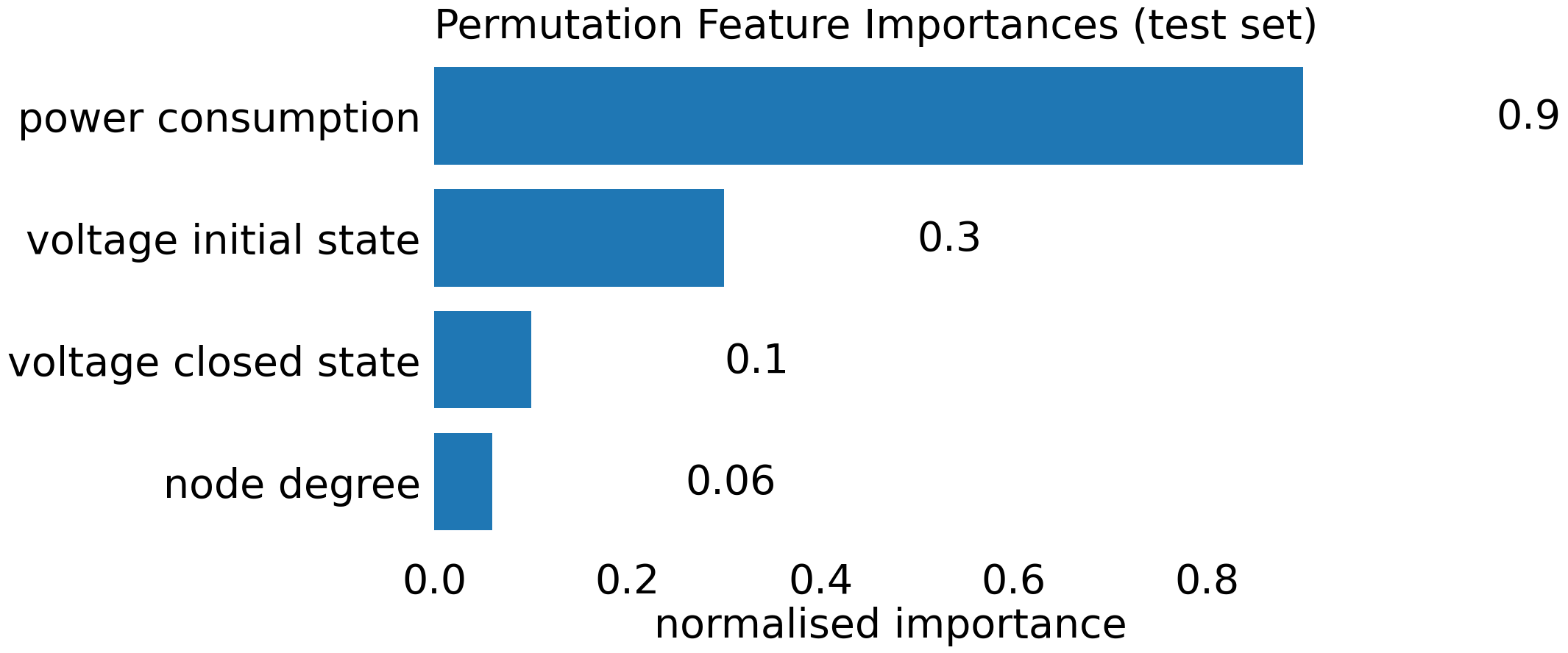}
         \caption{}
         \label{fig:perm_nodes}
     \end{subfigure}
     \hfill
     \begin{subfigure}[b]{3.5in}
         \centering
         \includegraphics[width=3.0in]{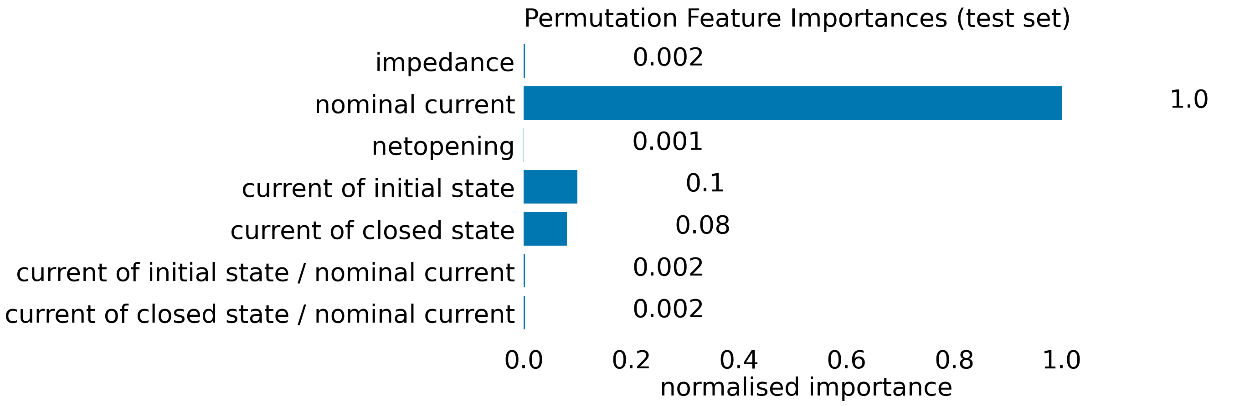}
         \caption{}
         \label{fig:perm_edges}
     \end{subfigure}
        \caption{a) Permutation Feature Importance for the node features, using the GIN-15 framework and test set. b) Permutation Feature Importance for the edge features, using the GIN-15 framework and test set.}
        \label{fig:PFI}
\end{figure*}

Permutation Feature Importance (PFI) \cite{fisher2019all} is a technique to determine the importance of features in predictive models. By systematically permuting the values of each feature and measuring the resulting impact on the predictive performance, PFI quantifies the relative significance of different features (nodes and edges). Figure \ref{fig:PFI} describes the PFI for the nodes and edges from a test set for assessing the predictions using the GIN-15 framework (global sum-pooling and augmented data). The results are averaged over all leave-one-out locations in Experiments III. Looking at the normalised importance, between 0 and 1, the power consumption (node feature) and the nominal current (edge feature) have the most influence as input data for predicting whether an input graph satisfies the n-1 principle. Regarding the node features, all show relevance to the prediction. The edge features, specifically the features related to current, show the most importance. For the application of the proposed approach in practice, the less important features can be reconsidered. In grids where there are voltage issues instead of current issues, other features can be more important.

\subsection{Comparison GNNs and mathematical optimisation approach}
Comparing the best-performing GIN framework (GIN-15) with the mathematical optimisation approach \cite{fritschy_h_checking_2018} shows a higher performance for the GIN in terms of accuracy (see Table \ref{tab:MILP_GNN}). Also, the mathematical optimisation approach requires significantly more computation time for n-1 evaluation because it searches for all switchover options on each grid. Note that, unlike machine learning models, the mathematical optimisation approach evaluates n-1 property per given sample, and there is no distinction between train/test data. On average, it uses 2300 seconds per sample which is equivalent to 38.3 minutes, and there are a total of 6200 samples in the test set resulting in 3961 hours of computational time required to assess n-1 property in all test samples. The training time of the GIN-15 model is $\sim$8.1 hours. The prediction time for the GIN-15 model is approximately 2.1 seconds per sample. The GIN-15 model is able to make much faster and more reliable assessments of the reliability of the grids compared to the mathematical optimisation approach.

\begin{table*}[]
\captionsetup{font=small}
\small
\centering
\caption{Comparisons of graph classification performance of GIN in terms of accuracy and the mathematical optimisation approach (with the time limit set to 2300 seconds per sample) on Experiments I (single locations). Best accuracy is highlighted in bold.}
\label{tab:MILP_GNN}
\begin{tabular}{l|llll|ll}
\hline
 & \multicolumn{4}{c|}{Experiments I}  &  &  \\ 
\multirow{-2}{*}{Model} & \multicolumn{1}{c}{1} & \multicolumn{1}{c}{2}  & \multicolumn{1}{c}{3}  & \multicolumn{1}{c !{\color{gray}\vrule}}{4} & \multirow{-2}{*}{\begin{tabular}[c]{@{}l@{}}Data \\ Augmentation\end{tabular}} & \multirow{-2}{*}{\begin{tabular}[c]{@{}l@{}}Global \\ Pooling\end{tabular}} \\ \hline
GIN-15   & \textbf{0.96  }&  \textbf{0.97} &  \textbf{0.97} &  \textbf{0.97}  & True & Sum\\
Mathematical Optimisation  &  0.90 & 0.92  & 0.93  &  0.88 & -- & --      \\       \hline
\end{tabular}
\end{table*}

\section{Conclusion}
Current algorithms are too slow to cope with the full complexity of energy grids. This paper proposes a GIN framework to assess the reliability of the medium voltage energy grid. Given the current energy transition, developing a GIN method that is robust to changing conditions (e.g., load/topology change) in unseen locations is of importance. The performance of the proposed method is verified on a dataset obtained from Alliander (DSO) and extended with additional augmented data.

The proposed approach demonstrates high accuracy, AUC and computational efficiency when applied to different operating locations and compared to the mathematical optimisation approach. Although the proposed GIN framework performs well in the selected locations and leave-one-out experiments (e.g., multi-source domain to single target domain), it is limited to large-scale grid regimes. The knowledge learned from the large-scale grids may not be sufficient for the model to predict the target task in sparse-grid regimes.  
Overall, we demonstrate that GINs have significant potential for improving the management and operation of the energy grid.

Non-i.i.d. graphs can be a limitation when having data with similar grid structures \cite{zhang_fairness_2023}. However, we do show that our approach generalises to unseen sparse grids. Regarding the structure of the grid, the Dutch energy grid is relatively comparable in different locations. For future research, it would be interesting to study the performance of GIN on even more diverse grid structures, for example, considering grids from different countries. Future research for the GNNs can be done in terms of edge/node-level classification (number of switchovers), multi-class classification (n-k principle), and transfer learning (data simulation of subgraphs). Furthermore, uncertainty quantification can be explored to support decision-making in practice.  

\bibliographystyle{IEEEtran}
\bibliography{references_doi}

\clearpage
\newpage 
\onecolumn
\appendix

\subsection{Accuracy}

\begin{table*}[h]
\captionsetup{font=small}
\small
\centering
\caption{Comparisons of graph classification performance of GIN in terms of accuracy of different settings on three experiment sets (single locations, all locations and unseen grids (leave-one-out (LOO))). Here, we implement four versions of GIN with different amounts of layers, three with different global pooling functions, and a version without data augmentation. Best accuracy is highlighted in bold.}
\label{tab:accuracy}
\begin{tabular}{l|llll!{\color{gray}\vrule}l!{\color{gray}\vrule}llll|ll}
\hline
 & \multicolumn{4}{c!{\color{gray}\vrule}}{Experiments I} & \multicolumn{1}{c!{\color{gray}\vrule}}{Experiments II} & \multicolumn{4}{c|}{Experiments III} &  &  \\ 
\multirow{-2}{*}{Model} & \multicolumn{1}{c}{1} & \multicolumn{1}{c}{2}  & \multicolumn{1}{c}{3}  & \multicolumn{1}{c !{\color{gray}\vrule}}{4} & \multicolumn{1}{c!{\color{gray}\vrule}}{1-4} & \multicolumn{1}{c}{LOO 1} & \multicolumn{1}{c}{LOO 2}  & \multicolumn{1}{c}{LOO 3}  & \multicolumn{1}{c|}{LOO 4} & \multirow{-2}{*}{\begin{tabular}[c]{@{}l@{}}Data \\ Augmentation\end{tabular}} & \multirow{-2}{*}{\begin{tabular}[c]{@{}l@{}}Global \\ Pooling\end{tabular}} \\ \hline

        GIN-5 &0.70 & 0.76 & 0.75 & 0.69 & 0.78 & 0.59 & 0.54 & 0.52 & 0.57  & True & Sum \\ 
        \rowcolor[HTML]{EFEFEF} GIN-10 &0.82 & 0.85 & 0.88 & 0.84 & 0.92 & 0.77 & 0.73 & 0.72 & 0.69 & True & Sum \\ 
        GIN-15 &\textbf{0.96} & \textbf{0.97} & \textbf{0.97} & \textbf{0.97} & \textbf{0.98} & \textbf{0.91} & \textbf{0.87} & \textbf{0.86 }& \textbf{0.90} & True & Sum \\ 
        \rowcolor[HTML]{EFEFEF} GIN-20 &0.94 & 0.95 & 0.95 & 0.96 & 0.96 & 0.90 & 0.86 & 0.85 & 0.88 & True & Sum \\ 
        GIN-15 &0.87 & 0.86 & 0.82 & 0.85 & 0.88 & 0.63 & 0.56 & 0.54 & 0.66 & False & Sum \\ 
        \rowcolor[HTML]{EFEFEF} GIN-15 &0.96 & 0.95 & 0.94 & 0.95 & 0.97 & 0.87 & 0.86 & 0.85 & 0.88 & True & Max \\ 
        GIN-15 &0.87 & 0.88 & 0.87 & 0.89 & 0.90 & 0.84 & 0.79 & 0.78 & 0.82 & True & Mean \\ \hline
    
\end{tabular}
\end{table*}

\end{document}